\documentclass[10pt,twocolumn,letterpaper]{article}

\usepackage{cvpr}              
\usepackage{multirow}
\usepackage{algorithm}
\usepackage{algorithmic}
\usepackage{amsfonts,amssymb}
\usepackage{booktabs}
\usepackage{comment}

\usepackage{colortbl}
\usepackage{color}

\definecolor{mygray}{gray}{.9}
\definecolor{light-gray}{gray}{0.5}
\definecolor{pretty-blue}{RGB}{0, 113, 188}
\definecolor{linecolor1}{gray}{.95} 
\definecolor{linecolor}{gray}{.895} 

\usepackage{bibentry}
\usepackage{multirow}
\usepackage{fancyhdr}
\usepackage{amsmath}
\usepackage{amssymb}
\usepackage{verbatim}
\usepackage{amsthm}

\newtheorem{theorem}{Theorem}
\newtheorem{corollary}[theorem]{Corollary}

\usepackage{newfloat}
\usepackage{listings}
\usepackage{graphicx} 

\definecolor{cvprblue}{rgb}{0.21,0.49,0.74}
\usepackage[pagebackref,breaklinks,colorlinks,allcolors=cvprblue]{hyperref}

\usepackage[capitalize]{cleveref}
\Crefname{section}{Section}{Sections}
\Crefname{table}{Table}{Tables}
\crefname{figure}{Figure}{Figures}
\crefname{equation}{Equation}{Equations}

\def\paperID{10745} 
\def\confName{CVPR}
\def\confYear{2025}

\title{Efficient Diffusion as Low Light Enhancer}

\author{
   Guanzhou Lan\textsuperscript{\rm 1,3}\footnotemark[1] \enspace
   Qianli Ma\textsuperscript{\rm 2}\footnotemark[1] \enspace
    Yuqi Yang\textsuperscript{\rm 1} \enspace \\
    Zhigang Wang\textsuperscript{\rm 1,3} \enspace
    Dong Wang\textsuperscript{\rm 3} \enspace
    Xuelong Li\textsuperscript{\rm 1,4} \enspace
    Bin Zhao\textsuperscript{\rm 1,3\footnotemark[2]} \\
    \textsuperscript{1}Northwestern Polytechnical University
  \enspace
    \textsuperscript{2}Shanghai Jiao Tong University  
  \enspace
    \textsuperscript{3}Shanghai AI Lab 
     \enspace
     \textsuperscript{4}TeleAI  \\
}

\begin{document}
\maketitle


{
\renewcommand{\thefootnote}{\fnsymbol{footnote}}
\footnotetext[1]{Equal Contribution.}
}

{
\renewcommand{\thefootnote}{\fnsymbol{footnote}}
\footnotetext[2]{Corresponding author.}
}

\begin{abstract}

The computational burden of the iterative sampling process remains a major challenge in diffusion-based Low-Light Image Enhancement (LLIE). Current acceleration methods, whether training-based or training-free, often lead to significant performance degradation, highlighting the trade-off between performance and efficiency.
In this paper, we identify two primary factors contributing to performance degradation: fitting errors and the inference gap. Our key insight is that fitting errors can be mitigated by linearly extrapolating the incorrect score functions, while the inference gap can be reduced by shifting the Gaussian flow to a reflectance-aware residual space.
Based on the above insights, we design Reflectance-Aware Trajectory Refinement (RATR) module, a simple yet effective module to refine the teacher trajectory using the reflectance component of images. Following this, we introduce \textbf{Re}flectance-aware \textbf{D}iffusion with \textbf{Di}stilled \textbf{T}rajectory (\textbf{ReDDiT}), an efficient and flexible distillation framework tailored for LLIE. Our framework achieves comparable performance to previous diffusion-based methods with redundant steps in just 2 steps while establishing new state-of-the-art (SOTA) results with 8 or 4 steps. Comprehensive experimental evaluations on 10 benchmark datasets validate the effectiveness of our method, consistently outperforming existing SOTA methods. Code is available at \href{https://mqleet.github.io/ReDDiT_Project/}{project page}.


%
%
\end{abstract}    
\section{Introduction}
\label{sec:intro}



%



\begin{figure}[t] 
\centering
\includegraphics[width=\linewidth]{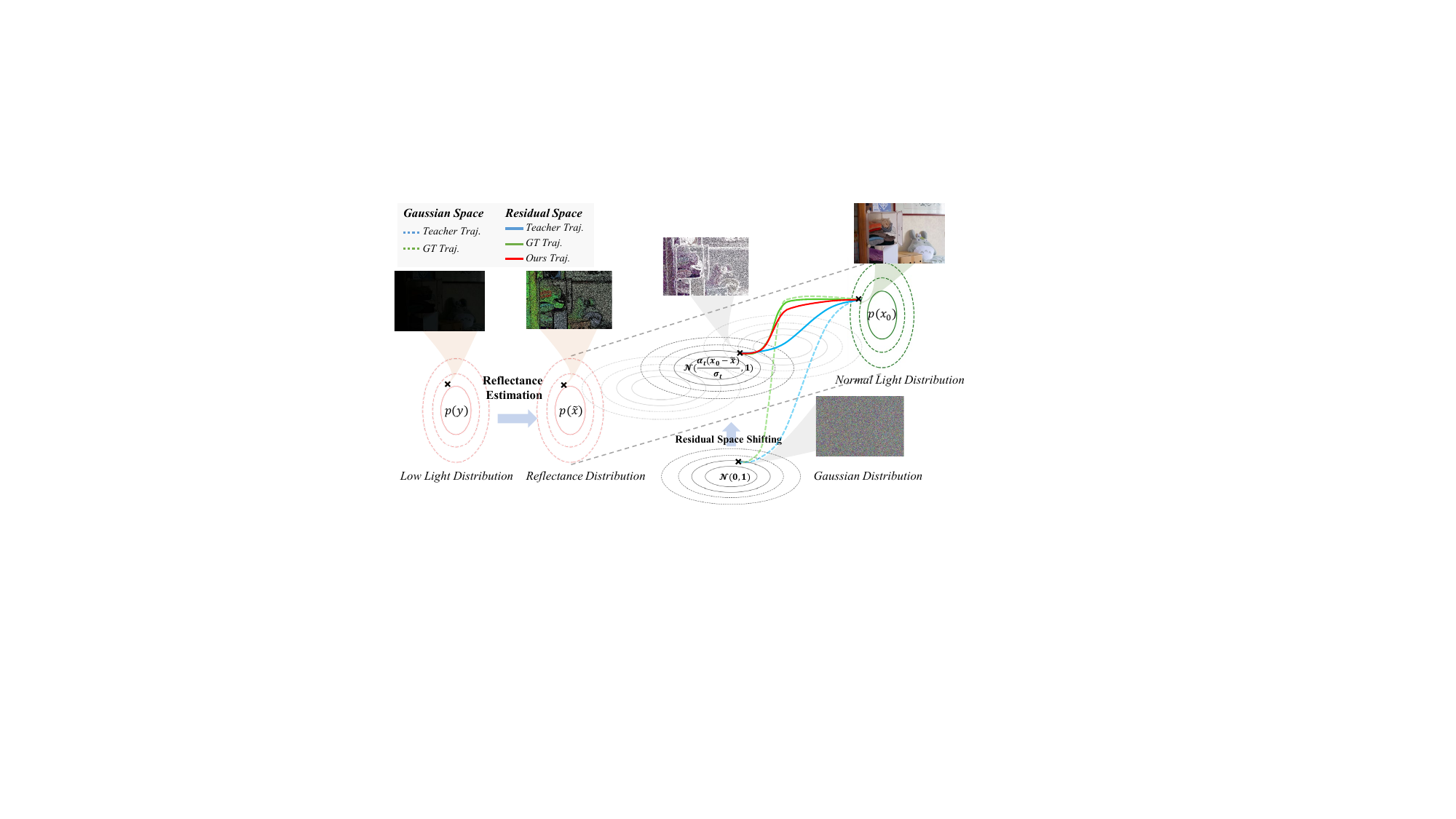} 
\vspace{-0.6cm}
 \caption{ReDDiT shifts the teacher trajectory from the original Gaussian distribution to a residual space, effectively reducing the sampling gap. Subsequently, it refines the teacher trajectory towards the ground truth trajectory to mitigate fitting errors. }
\vspace{-0.6cm}
\label{fig:top1}
\end{figure}

Low-light Image Enhancement (LLIE) is designed to improve the visibility and contrast of images captured in low-light conditions while maintaining natural-looking details, which is crucial for various downstream applications. Diffusion models~\cite{ddpm,rombach2022high} have achieved remarkable success in this domain, demonstrating significant progress in generating photorealistic normal light images~\cite{Diff-Retinex,yin2023cle,hou2024global}.
Previous diffusion-based LLIE methods~\cite{Diff-Retinex,yin2023cle,hou2024global} primarily focus on how to condition the low-light image within the diffusion framework. Diff-Retinex~\cite{Diff-Retinex} applies the Retinex decomposition to the low-light image and condition these components into the diffusion model. PyDiff~\cite{zhou2023pyramid} utilizes pyramid low-light images as conditions and applies DDIM~\cite{ddim} to expedite the sampling stage. In WCDM~\cite{diffll}, the diffusion model operates in the wavelet-transformed space, restoring the high-frequency parts of low-light images. GSAD~\cite{hou2024global} proposes a global structural regularization to enhance structural information learning, enabling smaller noise schedules for the inference stage.

However, the primary challenge to the broader application of diffusion models in LLIE stems from their iterative denoising mechanism. For example, DDPM~\cite{ddpm} requires multiple denoising steps, \eg,  1000 steps, to reverse a Gaussian noise into a clean image. The computation overhead conflicts with the demands of LLIE, particularly in computational photography applications for edge devices like mobile phones and surveillance cameras. 

Inspired by training-aware acceleration methods for diffusion models~\cite{salimans2021progressive, song2023consistency}, we investigate distilling a multi-step diffusion-based LLIE model to improve efficiency. A key observation is that performance degradation is inevitable as the number of sampling steps is reduced, even when using advanced acceleration methods (\eg, consistency distillation~\cite{song2023consistency}). This leads us to pose the question: \emph{Is it possible to distill a student diffusion model that surpasses the original diffusion model?} If so, even as performance degrades with fewer sampling steps, we could still achieve results comparable to those of the multi-step teacher model.

To address this, we conduct a comprehensive analysis of diffusion model acceleration techniques. Through theoretical analysis, we identify two primary factors contributing to performance degradation: \textbf{(1) The fitting error.}~It is the unavoidable errors between deep learning models and target fitting data. It will result the additional undesired terms and mismatch during distillation. \textbf{(2) The inference gap.}~It is the gap between the training target and sampling strategies specialized for the Diffusion models. It is caused by the universal diffusion model often trained and operated on the Gaussian flow for the diversity of generation, while the LLIE requires for more deterministic.

Our key insight is to refine the teacher trajectory by applying linear extrapolation to the score functions of the teacher model, thereby mitigating the adverse effects of fitting errors. Meanwhile, shifting the sampling trajectory to a deterministic space addresses the sampling gap, as shown in \cref{fig:top1}. A detailed theoretical analysis of this approach is presented in ~\cref{sec:method}.

Based on the above principles, we design Reflectance-Aware Trajectory Refinement (RATR) module, a simple yet effective module to refine the teacher trajectory for LLIE task. It incorporates the reflectance component of images as a deterministic prior to adjust and refine diffusion trajectories. Following this, we introduce \textbf{Re}flectance-aware \textbf{D}iffusion with \textbf{Di}stilled \textbf{T}rajectory (\textbf{ReDDiT}), an efficient and flexible distillation framework tailored for LLIE. This framework perform trajectory matching distillation is conducted between the refined teacher and student models, yields a 2-step diffusion model with comparable performance to previous 10-step diffusion-based methods, as well as 4 and 8-step diffusion models that achieve new SOTA results. our contributions are summarized as follows:

(1) We theoretically analyze the factors contributing to performance degradation and propose targeted improvements: linear extrapolation of the score function to mitigate performance loss due to fitting errors, and residual shifting as a solution to the sampling gap. 

(2) Based on these two design principles, we present ReDDiT, a novel distillation scheme that enhances the efficiency of generative diffusion models for LLIE. Notably, ReDDiT achieves high-quality image restoration in just 2 steps, establishing a new benchmark for efficient diffusion-based models in this domain.

(3) Extensive experiments conducted on 10 benchmark datasets validate that ReDDiT consistently achieves SOTA results, showcasing its superiority in terms of both quality and efficiency, even with a minimal number of steps.

\section{Related Work}

\textbf{Low-Light Image Enhancement.} Enhancing low-light images is a classical task in low-level vision, with numerous solutions leveraging deep neural networks~\cite{guo2020zero, jiang2021enlightengan, Wu_2022_CVPR}. Recently, as the diffusion model has demonstrated promising results in image generation tasks, there has been growing interest in leveraging diffusion models for LLIE to achieve faithful restoration results. Diff-Retinex applies Retinex decomposition as the condition of diffusion model~\cite{Diff-Retinex}. PyDiff utilizes pyramid low-light images as the condition and applies DDIM to speed up the sampling stage~\cite{zhou2023pyramid, ddim}. In WCDM, the diffusion model is constructed in the wavelet-transformed space and restores the high-frequency parts of low-light images~\cite{diffll}. GSAD proposes a global structural regularization to enhance structural information learning, which also helps to train diffusion models with less curvature, enabling smaller noise schedules for the inference stage~\cite{hou2024global}. Unfortunately, these works only consider how to condition the low-light images, neglecting the efficiency concerns.

\noindent\textbf{Diffusion Model Acceleration.} Diffusion models have shown promising performance across various tasks such as style transfer~\cite{ControlStyle}, video generation~\cite{MV-Diffusion}, and text-to-image generation~\cite{li2024snapfusion,ruiz2023dreambooth,rombach2022high}. However, the issue of redundant sampling steps remains a significant efficiency concern. Recent research has explored various strategies to accelerate diffusion models.
One series of methods involves the use of fast post-hoc samplers~\cite{lu2022dpm,lu2022fast,ddim}, which reduces the number of inference steps for pre-trained diffusion models to 20-50 steps. However, most suffer from severe performance degradation when further accelerating sampling within 10 steps. To address this limitation, step distillation~\cite{luhman2021knowledge,salimans2021progressive} is proposed, aiming to distill diffusion models into fewer steps. Progressive distillation (PD)~\cite{salimans2021progressive} is the first successful practice and produces 2-step unconditional diffusion. Following PD~\cite{salimans2021progressive}, \citeauthor{meng2023distillation} apply PD to the large-scale Stable Diffusion, achieving text-to-image generation with 2 steps. Consistency Distillation (CD)~\cite{song2023consistency} aims to learn consistency among diffusion timesteps of the teacher models~\cite{song2023consistency}.
Following this line of research, CTM~\cite{CTM} extends such consistency from individual timesteps to an entire trajectory along the diffusion model, enabling faithful unconditional generation. Despite the booming development of diffusion distillation in other fields, techniques for LLIE are still left blank. In this paper, we present the first distilled diffusion tailored for LLIE, which will be introduced in detail in the following sections.

\begin{figure*}[t]
\centering
\includegraphics[width=0.9\linewidth]{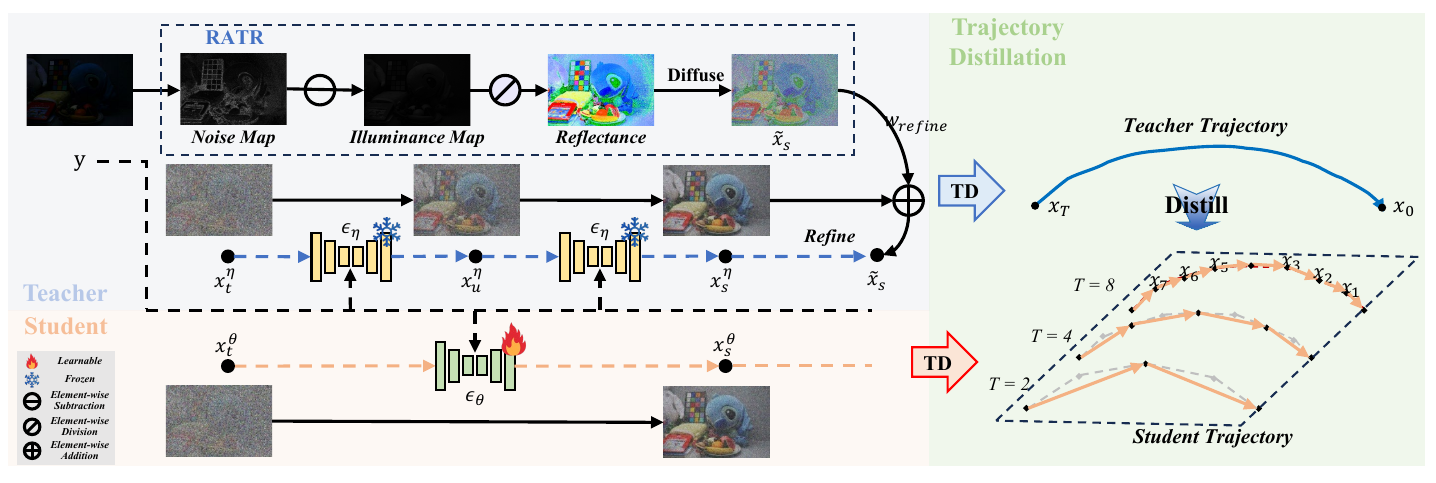} 
\vspace{-0.3cm}
\caption{Pipeline of our proposed ReDDiT. The distillation process involves two parts: teacher model leverages the estimated reflectance to refine its trajectory and student model's trajectory is guided by teacher's trajectory, via a distillation loss. \textbf{TD} denotes the Trajectory Decoder while \textcolor{blue}{\textbf{RATR}} denotes the Reflectance-Aware Trajectory Refinement.}
\vspace{-0.5cm}
\label{fig:main}
\end{figure*}

\section{Methods}
\label{sec:method}

Let $x_{0}$ denote a high-quality image, the forward diffusion process aims to generate a sequence of noisy latent variables $x_1, x_2, ..., x_T$ using a Markovian process, which is defined by the equation:
\begin{equation}
    \begin{aligned}
        x_t = \alpha_t x_0 + \sigma_t\epsilon ,
    \end{aligned}	
    \label{eq: c-ddpm-forward}
\end{equation}
where $ \alpha_t \in(0,1)$ represents the noise schedule, $\sigma_t$ denotes the covariance at $t$, and $\epsilon$ represents the Gaussian noise. In the reverse process, we incorporate the low-light image $y$ as the conditioning of the score functions $\epsilon_\eta(x_t, y, \alpha_t)$ and predict the clean images iteratively:
\begin{equation}
    \begin{aligned}
      x^\eta_{t-1}= \alpha_{t-1} \left(\frac{x_t - \sigma_t\epsilon_\eta}{\alpha_t} \right) + \sigma_{t-1}\epsilon_\eta .
    \end{aligned}	
    \label{eq: c-ddpm-reserved}
\end{equation}

In this section,  we propose ReDDiT, aiming to train a student diffusion model with parameters $\theta$ to learn the trajectory of the teacher model $\eta$ in fewer steps. We begin by introducing trajectory distillation. Next, we theoretically analyze the factors contributing to performance degradation and present a concise formulation for refining the trajectory. Then, we determine that the reflectance component is most effective for refining the teacher trajectory and design a Reflectance-Aware Trajectory Refinement (RATR) module. Finally, all components are integrated into the ReDDiT framework, where distillation is performed.

\subsection{Trajectory Distillation}
 
 The essence of diffusion distillation differs from traditional knowledge distillation is that distilled diffusion models learn the trajectory from the teacher, preserving the iterative sampling characteristics unique to diffusion models. We define $G(x_t, y, t, s)$ as the trajectory decoder that transition from time step $t$ to $s$ $(s<t)$ to represent the diffusion model trajectory. Draw inspiration from the denoising diffusion implicit model, we represent the decoder with the student score function $\epsilon_\theta$ as:
\begin{equation}
    \begin{aligned}
    G_{\theta}(x_t, y, t, s)  =\frac{\alpha_s}{\alpha_t}x_t + (\sigma_s - \frac{\alpha_s}{\alpha_t} \sigma_t) \epsilon_\theta(x_{t}, y, t) .
    \end{aligned}
    \label{eq: student_traj}
\end{equation}
With this trajectory decoder, the teacher model's trajectory can be decoded from time step $s$ to $t$. For a more precise estimation of the entire teacher trajectory, we also utilize the intermediate step $u \in [s, t)$ to estimate the trajectory, which formulates the trajectory decoder in the second order as: 
\begin{equation}
    \begin{aligned}
    x^\eta_{s,u,t} & = G_\eta(G_\eta(x_t, y, t, u), y, u, s) \\
        & =  \frac{\alpha_s}{\alpha_t}x_t + \sigma_s\epsilon_\eta(x_{u}, y, u) \\ 
    &+ \frac{\alpha_s}{\alpha_u} \sigma_u (\epsilon_\eta(x_{t}, y, t) - \epsilon_\eta(x_{u}, y, u)) \\
      & - \frac{\alpha_s}{\alpha_t} \sigma_t\epsilon_\eta(x_{t}, y, t),
    \end{aligned}
    \label{eq: teacher_traj}
\end{equation}
To facilitate trajectory learning during distillation, the student model should match the teacher trajectory from $t$ to $s$. Denoting the student trajectory as $x^\theta_{s,t} = G_\eta(x_t, y, t, s)$, the distillation regularization is formulated as:
 \begin{equation}
    \begin{aligned}
    \theta= \arg\min_\theta \| x^\theta_{s,t} -  x^\eta_{s,u,t} \|_2^2 .
    \end{aligned}
\end{equation}
With such regularization, the information intermediate step $u \in [s, t)$ will be distilled in student model. In practice, we utilize the clean image predicted by the trajectory both on the student and teacher to perform distillation for stable training. The predicted clean images of the teacher and student, denoted as $ x_{target}$ and $x_{est}$, are formulated as follows:
\begin{equation}
    \begin{aligned}
      x_{target} & =\frac{ x^\eta_{s,u,t}-(\sigma_{s}/\sigma_{t})x_{t}}{\bar{\alpha}_{s}-(\sigma_{s}/\sigma_{t})\bar{\alpha}_{t}}, \quad x_{est}  = \frac{x^\theta_{s,t}-(\sigma_{s}/\sigma_{t})x_{t}}{\bar{\alpha}_{s}-(\sigma_{s}/\sigma_{t})\bar{\alpha}_{t}} ,
    \end{aligned}	
\end{equation}
and the distillation regularization is modified as:
\begin{equation}
    \begin{aligned}
  \theta= \arg\min_\theta\lambda(t) \|x_{target}-x_{est}\|_2^2 , \\
    \end{aligned}
\end{equation}
where $\lambda(t)$ is an adaptive weight and is set as $ \max(1, \frac{\alpha_t^2}{\sigma_t^2})$.

\subsection{On the Refinement of the Teacher Trajectory}
Direct application of trajectory distillation often leads to significant performance degradation. In this section, we theoretically analyze the core reasons behind this degradation and propose a strategy to mitigate its effects.

\noindent \textbf{On the fitting error of the teacher trajectory.}
To understand this, let us revisit \cref{eq: student_traj} and \cref{eq: teacher_traj}, and consider the ideal condition where $\mathcal{L}_{distill}=0$, meaning that the student’s trajectory perfectly matches the teacher’s trajectory. Under this condition, we have:
\begin{equation}
    \begin{aligned}
   (\sigma_s - \frac{\alpha_s}{\alpha_t} \sigma_t) \epsilon_\theta(x_{t}, y, t) 
   & =    \sigma_s\epsilon_\eta(x_{u}, y, u)  - \frac{\alpha_s}{\alpha_t} \sigma_t\epsilon_\eta(x_{t}, y, t) \\ 
    & + \frac{\alpha_s}{\alpha_u} \sigma_u \Big( \epsilon_\eta(x_{t}, y, t) - \epsilon_\eta(x_{u}, y, u) \Big).  \\
    \end{aligned}
    \label{eq: traj_match}
\end{equation}

Since the teacher model is trained using the vanilla diffusion loss function, the ideal teacher trajectory satisfies the condition $\epsilon_\eta(x_{t}, y, t)=\epsilon_\eta(x_{u}, y, u) = \tilde{\epsilon}$. Under this condition, the \cref{eq: traj_match} can be simplified as:
\begin{equation}
    \begin{aligned}
   (\sigma_s - \frac{\alpha_s}{\alpha_t} \sigma_t) \epsilon_\theta(x_{t}, y, t) & =    (\sigma_s - \frac{\alpha_s}{\alpha_t} \sigma_t) \tilde{\epsilon}.\\ 
    \end{aligned}
    \label{eq: traj_match_simplify}
\end{equation}
However, the existence of fitting errors makes it impossible to achieve this goal. The presence of guidance with undesired components inevitably leads to suboptimal results.

\noindent \textbf{On the mitigation of the fitting error.}
Fortunately, we find that these undesired terms can be mitigated by a scaling parameter $\omega \in (0,1]$. For the term $\frac{\alpha_s}{\alpha_u} \sigma_u \Big( \epsilon_\eta(x_{t}, y, t) - \epsilon_\eta(x_{u}, y, u) \Big) $, we can further reduce its impact by multiplying it by the scaling parameter $\omega$. The mismatch between $\epsilon_\eta(x_{u})$ and $\epsilon_\eta(x_{t})$ introduces a more significant optimization conflict. Due to the presence of fitting error, this term is not always consistent with $\epsilon_\eta(x_{t}, y, t) $. Our key insight is to refine this term through linear interpolation toward the ideal value, expressed as $\sigma_t \Big(\omega \epsilon_\eta(x_{t}, y, t) +  (1-\omega)\tilde{\epsilon} \Big)$. This approach not only maintains consistency with previous terms, but also provides linear trajectory guidance. After applying these operations, the refined teacher trajectory can be expressed as follows:
\begin{equation}
    \begin{aligned}
   \tilde{x}^{\eta}_{s,u,t} & =  \frac{\alpha_s}{\alpha_t}x_t + \sigma_s\epsilon_\eta(x_{u}, y, u) \\ 
    &+ \omega  \frac{\alpha_s}{\alpha_u} \sigma_u \Big(\epsilon_\eta(x_{t}, y, t) - \epsilon_\eta(x_{u}, y, u)\Big) \\
      & - \frac{\alpha_s}{\alpha_t} \sigma_t \Big(\omega \epsilon_\eta(x_{t}, y, t) +  (1-\omega)\tilde{\epsilon}\Big) .\\
    \end{aligned}
    \label{eq: teacher_traj_re}
\end{equation}
To this end, the distillation loss, $\mathcal{L}_{distill}$, is equivalent to the following, disregarding the coefficients of each term:
\begin{equation}
    \begin{aligned}
   \mathcal{L}_{distill}& = ||\epsilon_\theta(x_{t}, y, t) - \epsilon_\eta(x_{u}, y, u) || \\
   &+ ||\epsilon_\theta(x_{t}, y, t) - (\omega \epsilon_\eta(x_{t}, y, t) +  (1-\omega)\tilde{\epsilon}) || .\\
    \end{aligned}
    \label{eq: distill_re}
\end{equation}

\noindent \textbf{On the mitigation  of the inference gap.}
Considering the vanilla diffusion are trained on the Gaussian flow, we determine the $\tilde{\epsilon}$ not the pure Gaussian noise but shift it into a residual space as:
\begin{equation}
    \begin{aligned}
    \tilde{\epsilon}= \frac{x_t-\alpha_t\tilde{x}_0}{\sigma_t} = \frac{\alpha_t(x_0-\tilde{x}_0)}{\sigma_t} + \epsilon ,\\
    \end{aligned}
    \label{eq: deterministic_shifting}
\end{equation} 
where  $\tilde{{x}}_0$ should lie between the low-light and clean image spaces, serving as an intermediate initial space for the student model to learn. This positioning ensures a closer initial distribution for the student model compared to the Gaussian distribution.

Additionally, our investigation shows that the refinement of the teacher trajectory in \cref{eq: teacher_traj_re} can be implemented through a straightforward approach, as outlined in the Corollary~\ref{cor:refinement}.


\begin{corollary}[Proof in the supplementary material]
\label{cor:refinement}
Given the refinement component $\tilde{x}_s = \alpha_s \tilde{x}_0 + \sigma_s\epsilon_\eta $,  the \cref{eq: teacher_traj_re} is equivalent to :
    $\tilde{x}^{\eta}_{s,u,t}  = \omega x^{\eta}_{s,u,t} + (1 - \omega) \tilde{x}_s $.
\end{corollary}


\subsection{Reflectance-Aware Trajectory Refinement}
On the determination of the component $\tilde{x}_0$, using the ground truth clean images $x_0$ will cause $\tilde{\epsilon}$ degrade to $\epsilon$. A natural approach is to use the low-light images $ y $ as $\tilde{x}_0$. In practice, the reflectance, which shares characteristics with both the clean images and the low-light images $y$, serves as a better component for refining the trajectory. Building on this observation, we propose the RATR module to refine the teacher's trajectory, thereby reducing inference gap.

Given an illumination map $h$ and the noise $z$, the reflectance can be obtained through $x = \frac{y -z}{h}$ based on the Retinex theory.  For illumination estimation, we employ the maximum channel of the low-light image $y$ to represent the estimated illumination map $h^\prime$, as the common practice. Regarding ISO noise estimation, similar to the previous non-learning-based denoising method \cite{buades2005non}, the noise can be modeled as the distance between the noisy image and the clean image. We use this distance to estimate the noise map of the input low-light images:
\begin{equation}
    \begin{aligned}
     z^\prime= |y-\psi(y)|,
    \end{aligned}	
\end{equation}
where $\psi$ represents a non-learning-based denoising operation, allowing for flexible distillation. With these estimations, we can obtain a latent clean image $\tilde{x}_0 = \frac{y - z^\prime}{h^\prime}$. The $\tilde{x}_0$ is then employed to the trajectory refinement to perform distillation. Extensive experiments in ablation studies show that such refinement reduce the inference gap and greatly improves performance. 


\begin{figure*}[ht] 
\centering
\includegraphics[width=0.9\linewidth]{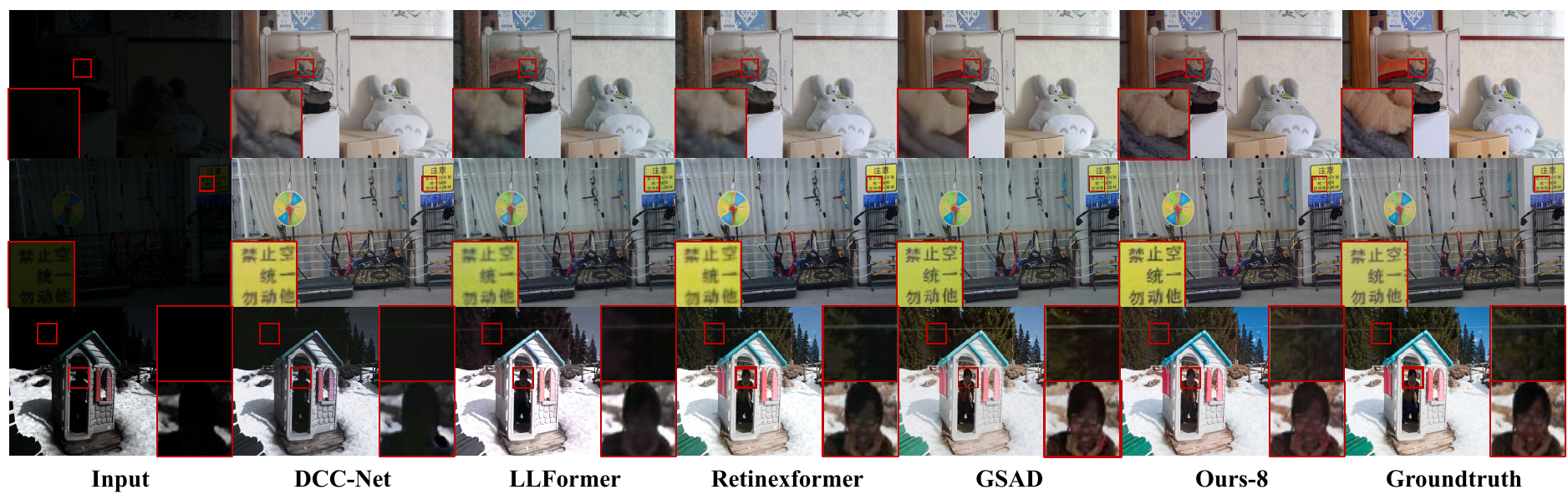} 
\vspace{-0.2cm}
 \caption{Qualitative results on LOLv1 (top) , LOLv2-real (middle), and LOLv2-synthetic (bottom). Patches highlighted in each image by the red box indicate that ReDDiT effectively enhances the visibility, preserves the color and generates finer details in normal light images. Zoom in to better observe the image details.}
\vspace{-0.4cm}
\label{fig:visual_compare}
\end{figure*}


\subsection{Auxiliary Loss }
In knowledge distillation for classification, direct training signal derived from the data label will help student classifier outperforms the teacher classifier.  
In ReDDiT, we extend the principles to our model by introducing direct signals from both pixel and feature spaces. In the pixel space, we employ L2 loss on the pixel space:
\begin{equation}
    \begin{aligned}
    \mathcal{L}_{pix} = \lambda_{pix} \|x_0-x_{est}\|_2^2 .
    \end{aligned}
\end{equation}
In the feature space, we employ the perceptual loss for enhancing the student learning in structure and texture details:
\begin{equation}
    \begin{aligned}
    \mathcal{L}_{per}&= \lambda_{per} \| \phi(x_0)-\phi(x_{est})\|_2^2.\\
    \end{aligned}
\end{equation}
The final loss in the distillation stage integrates all the components into a single and unified training framework:
\begin{equation}
    \begin{aligned}
    \mathcal{L}_{total}&=\mathcal{L}_{distill} +  \mathcal{L}_{per} + \mathcal{L}_{pix} .
    \end{aligned}
\end{equation}

\section{Experiments}
\begin{table*}[ht]
\centering
\resizebox{0.9\linewidth}{!}{
    \begin{tabular}{l | c |c  c  c |c  c  c | c  c  c   }
    \toprule \multirow{2}{*}{ \textbf{Methods} } & \multirow{2}{*}{ \textbf{NFE} } & \multicolumn{3}{c|}{ \textbf{LOLv1} } & \multicolumn{3}{c|}{ \textbf{LOLv2-real} } & \multicolumn{3}{c}{ \textbf{LOLv2-synthetic} }  \\ 
    & & PSNR  $\uparrow$  & SSIM  $\uparrow$  & LPIPS  $\downarrow$  & PSNR  $\uparrow$  & SSIM  $\uparrow$  & LPIPS  $\downarrow$  & PSNR  $\uparrow$  & SSIM  $\uparrow$  & LPIPS  $\downarrow$   \\
    \midrule LIME \cite{LIME} (TIP 16) & \multirow{11}{*}{ 1 } & 16.760 & 0.560 & 0.350 & 15.240 & 0.470 & 0.415 & 16.880 & 0.776 & 0.675 \\
    Zero-DCE \cite{guo2020zero}  (CVPR 20) & & 14.861 & 0.562 & 0.335 & 18.059 & 0.580 & 0.313 & - & - & -  \\
    EnlightenGAN \cite{jiang2021enlightengan} (TIP 21) & & 17.483 & 0.652 & 0.322 & 18.640 & 0.677 & 0.309 & 16.570 & 0.734 & -  \\
    RetinexNet \cite{lol1} (BMVC 18) & &16.770 & 0.462 & 0.474 & 18.371 & 0.723 & 0.365 & 17.130 & 0.798 & 0.754 \\
    DRBN \cite{DRBN} (CVPR 20) & &19.860 & 0.834 & 0.155 & 20.130 & 0.830 & 0.147 & 23.220 & 0.927 & -  \\
    KinD \cite{zhang2019kindling} (ACM MM 19) & &20.870 & 0.799 & 0.207 & 17.544 & 0.669 & 0.375 & 16.259 & 0.591 & 0.435 \\
    KinD++ \cite{zhang2019kindling} (IJCV 20) & &21.300 & 0.823 & 0.175 & 19.087 & 0.817 & 0.180 & - & - & -\\
    MIRNet \cite{mirnet} (TPAMI 22) & &24.140 & 0.842 & 0.131 & 20.357 & 0.782 & 0.317 & 21.940 & 0.846 & - \\
    LLFlow \cite{llflow} (AAAI 22) & &25.132 &  0.872  &  0.117  & 26.200 & 0.888  &  0.137  & 24.807 & 0.9193 & 0.067 \\
    SNR-Net \cite{xu2022snr} (CVPR 22) & &26.716 & 0.851 & 0.152 & 27.209 & 0.871 & 0.157 & 27.787 &  0.941  &  0.054  \\
    LLFormer \cite{llformer} (AAAI 23) & &25.758 & 0.823 & 0.167 & 26.197 & 0.819 & 0.209 & 28.006 & 0.927 & 0.061 \\
    Retinexformer \cite{cai2023retinexformer} (ICCV 23) & &27.180 & 0.850 & - & 27.710 & 0.856 & - & 29.040 & 0.939 & - \\
   
    \midrule 
    PyDiff \cite{zhou2023pyramid} (IJCAI 23) & 4 & 27.090  &  \bf{0.930}  &  0.100  &  29.629  &  0.876  & 0.194  &  20.659  &  0.879  &  0.275   \\
    WCDM \cite{diffll} (ToG 23) & 10 & 26.947  &  0.847  &  0.088  &  30.461  &  0.879  & 0.070  &  29.852 &  0.908  &  0.077\\
    GSAD \cite{hou2024global} (NIPS 23) & 10 & 27.839  &   \underline{0.877}  &  0.091  &  28.818  &  0.895  & 0.095  &  28.670  &  \bf{0.944}  &  0.047   \\
    Ours-8\cellcolor{linecolor1} & 8\cellcolor{linecolor1} & \bf{28.090}\cellcolor{linecolor1}  &  0.861\cellcolor{linecolor1}  &  \bf{0.052}\cellcolor{linecolor1}  & \underline{30.919}\cellcolor{linecolor1}  &  \underline{0.895}\cellcolor{linecolor1}  &\underline{0.040}\cellcolor{linecolor1}   &  \bf{30.166}\cellcolor{linecolor1}  &  \underline{0.943}\cellcolor{linecolor1}  &  \bf{0.028}\cellcolor{linecolor1}   \\
    Ours-4\cellcolor{linecolor1} & 4\cellcolor{linecolor1} & \underline{27.981}\cellcolor{linecolor1}  &  0.872\cellcolor{linecolor1}  &  \underline{0.053}\cellcolor{linecolor1}  &  \bf{31.250}\cellcolor{linecolor1}  &  \bf{0.904}\cellcolor{linecolor1}  & \bf{0.038}\cellcolor{linecolor1}  &  \underline{30.034}\cellcolor{linecolor1}  & 0.933\cellcolor{linecolor1}  &  \underline{0.031}\cellcolor{linecolor1}   \\
    Ours-2\cellcolor{linecolor1} & 2\cellcolor{linecolor1}& 27.397\cellcolor{linecolor1}  &  0.848\cellcolor{linecolor1}  &  0.062\cellcolor{linecolor1}  &  30.613\cellcolor{linecolor1}  &  0.890\cellcolor{linecolor1}  & 0.045\cellcolor{linecolor1}  &  29.346\cellcolor{linecolor1} & 0.913\cellcolor{linecolor1} &  0.040\cellcolor{linecolor1}   \\
    \bottomrule
    \end{tabular}}
     \vspace{-0.2cm}
  \caption{Quantitative comparisons are conducted on the LOLv1 and LOLv2 (real and synthetic) datasets. The best result is highlighted in bold, while the second-best result is underlined. Our method significantly outperformed SOTA methods. }
  \vspace{-0.4cm}
 \label{tab:LOL}
\end{table*}

\begin{figure*}[ht]
\centering
\includegraphics[width=0.9\linewidth]{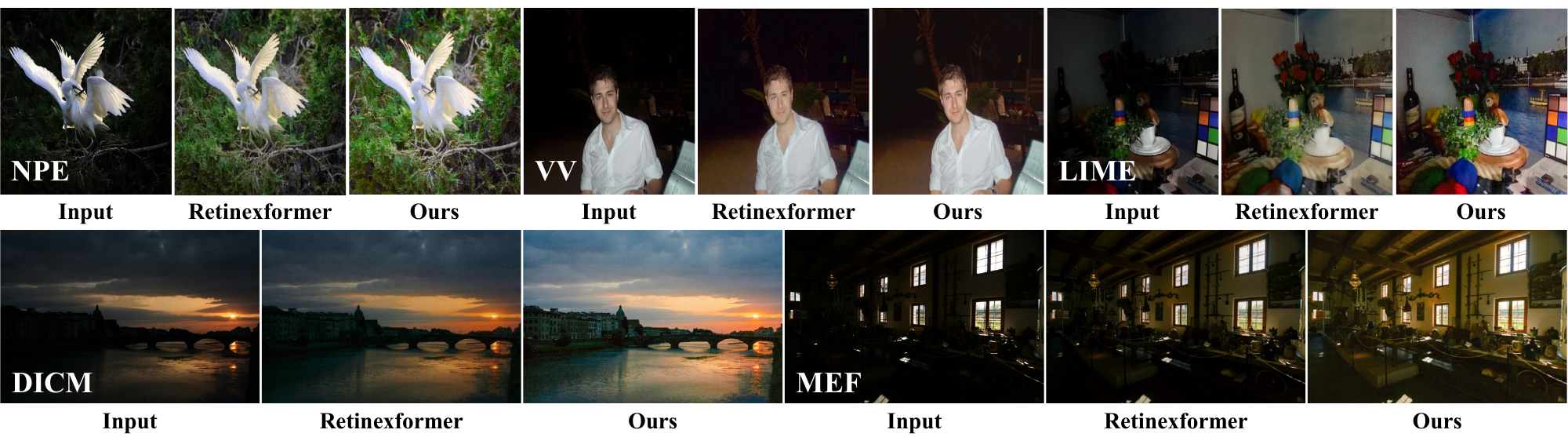} 
\vspace{-0.2cm}
 \caption{Qualitative results on NPE, VV, LIME, DICM, and MEF. Compared to Retinexformer, our method effectively enhances the visibility and preserves the color in normal light images. Zoom in to better observe the image details.}
\vspace{-0.2cm}
\label{fig:unpair}
\end{figure*}

\begin{table}[ht]
    \centering
    \resizebox{0.95\linewidth}{!}{ 
    \begin{tabular}{l|cc|cc}
    \toprule  
    \multirow{2}{*}{ \textbf{Methods}} & \multicolumn{2}{c|}{ \textbf{SID} } &  \multicolumn{2}{c}{ \textbf{SDSD} }   \\

    &PSNR $\uparrow$ & SSIM $\uparrow$ & PSNR $\uparrow$ & SSIM $\uparrow$ \\
    \midrule 
    SID \cite{sid} (CVPR 18) &16.97 & 0.591 & 23.29 & 0.703   \\
    RetinexNet \cite{lol1} (BMVC 18) &16.48 & 0.578 & 20.84 & 0.617   \\
     DeepUPE \cite{deepupe} (CVPR 19) &17.01 & 0.604 & 21.70 & 0.662   \\
    KinD \cite{zhang2019kindling} (ACM MM 19) &18.02 & 0.583 & 21.95 & 0.672  \\
    3DLUT \cite{3dLUT} (TPAMI 20) &20.11 & 0.592 & 21.66 & 0.655  \\
    DRBN \cite{DRBN} (CVPR 20) &19.02 & 0.577 & 24.08 & 0.868  \\
     EnGAN \cite{jiang2021enlightengan} (TIP 21) &17.23 & 0.543 & 20.02 & 0.604  \\
    Uformer \cite{wang2022uformer} (CVPR 22) &18.54 & 0.577 & 23.17 & 0.859   \\
    Restormer \cite{zamir2022restormer} (CVPR 22) &22.27 & 0.649 & 25.67 & 0.827   \\
     MIRNet \cite{mirnet} (TPAMI 22) &20.84 & 0.605 & 24.38 & 0.864   \\
    SNR-Net \cite{xu2022snr} (CVPR 22) &22.87 & 0.625 & 29.44 & \underline{0.894}  \\
    Retinexformer \cite{cai2023retinexformer} (ICCV 23) &24.44 & 0.680 & 29.77 & \bf{0.896}   \\
    Ours-8\cellcolor{linecolor1}  & 25.32\cellcolor{linecolor1}  & \bf{0.802}\cellcolor{linecolor1}  & \bf{29.95}\cellcolor{linecolor1}  & 0.883\cellcolor{linecolor1}     \\
    Ours-4\cellcolor{linecolor1}  &\underline{25.85}\cellcolor{linecolor1}  & \underline{0.792}\cellcolor{linecolor1}  & \underline{29.52}\cellcolor{linecolor1}  & 0.857\cellcolor{linecolor1}    \\
    Ours-2\cellcolor{linecolor1}  &\bf{25.88}\cellcolor{linecolor1}  & 0.745\cellcolor{linecolor1}  & 28.10\cellcolor{linecolor1}  & 0.731\cellcolor{linecolor1}    \\
    \bottomrule
    \end{tabular}
    }
    \caption{Quantitative comparisons on SID and SDSD datasets. The best result is
    highlighted in bold, while the second-best result is underlined. }
    \label{tab:SID_SDSD}
    \vspace{-0.5cm}
\end{table}

\subsection{Dataset and Implementation Details}
\textbf{Datasets.} We evaluate the performance of ReDDiT across various datasets in low-light image regions, including LOLv1 \cite{lol1}, LOLv2 \cite{lol2}, SID \cite{sid}, SDSD \cite{sdsd}, DICM \cite{DICM}, LIME \cite{LIME}, MEF \cite{MEF}, NPE \cite{NPE}, and VV \cite{VV}.  



\noindent \textbf{Evaluation metrics.}
We comprehensively evaluate various Low-Light Image Enhancement (LLIE) methods, employing both full-reference and non-reference image quality metrics. In cases with paired data, we measure the peak signal-to-noise ratio (PSNR) and structural similarity index (SSIM), along with the learned perceptual image patch similarity (LPIPS). For datasets such as DICM, LIME, MEF, NPE, and VV, which lack paired data, we rely on the Naturalness Image Quality Evaluator (NIQE) for assessment.

\begin{table}[ht]
    \centering
    \resizebox{0.85\linewidth}{!}{
    \begin{tabular}{l|ccccc}
    \toprule 
    \textbf{Methods} & \textbf{DICM} & \textbf{LIME} & \textbf{MEF} & \textbf{NPE} & \textbf{VV} \\
    \midrule 
    LIME \cite{LIME}  &5.42& 3.84   &3.73   &4.37   &3.63   \\
    Zero-DCE \cite{guo2020zero} & 3.77 & 3.61 & 4.12 & 6.19 & 4.81 \\
    EnGAN \cite{jiang2021enlightengan} & 3.98 & 4.49 & 4.95 & 3.42 & 3.93 \\
    KinD++  \cite{zhang2021beyond} & 6.19 & 4.24 & 3.93 & 3.44 & 3.72\\
    SCI \cite{sci}  &3.73  &3.52   & 4.07  &6.44  & 4.05  \\
    DCC-Net \cite{dcc-net} & 3.84  &6.27   &6.04  &7.97 &  4.66 \\
    Retinexformer \cite{cai2023retinexformer}  &4.47   & 3.96 & 4.12 & 4.28 &4.53   \\
    WCDM \cite{diffll} &6.38   & 4.62 & 4.86 & 4.65 & 5.39   \\
    GSAD \cite{hou2024global} & 4.35 & 3.97 &4.08 &5.35& 3.18 \\
    Ours\cellcolor{linecolor1}  & \bf{3.62}\cellcolor{linecolor1} & \bf{3.45}\cellcolor{linecolor1} & \bf{3.93}\cellcolor{linecolor1} & \bf{3.24}\cellcolor{linecolor1} & \bf{3.00}\cellcolor{linecolor1} \\
   
     \bottomrule
    \end{tabular}
    }
    \caption{Quantitative comparisons of different methods on DICM, LIME, MEF, NPE, and VV in terms of NIQE. Lower values indicate better performance, with the best result highlighted in bold.}
     \vspace{-0.4cm}
     \label{tab:unpaired}
\end{table}
 
\begin{figure*}[ht]
\centering
\includegraphics[width=1.0\linewidth]{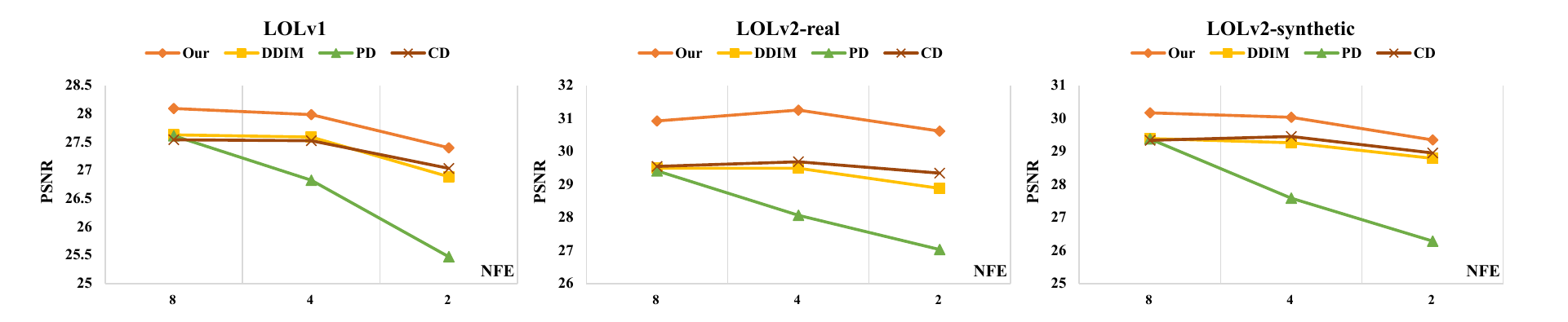} 
\vspace{-0.7cm}
 \caption{Quantitative comparisons on the LOLv1 and LOLv2 (Real and Synthetic) datasets with other accelerate methods. Popular acceleration methods experience a significant decline in performance as the number of steps decreases. Our method outperforms other acceleration techniques as well as \textbf{the original teacher model (in terms of DDIM performance)}.}
\vspace{-0.4cm}
\label{fig:comparison_with_acc}
\end{figure*}

\subsection{Comparison with State-of-the-Art Methods}
\textbf{Results on LOLv1 and LOLv2.} On LOLv1 and LOLv2, we compare ReDDiT against LIME \cite{LIME}, RetinexNet \cite{lol1}, KinD \cite{zhang2019kindling}, Zero-DCE \cite{guo2020zero}, DRBN \cite{DRBN}, KinD++ \cite{zhang2021beyond}, EnlightenGAN \cite{jiang2021enlightengan}, MIRNet \cite{mirnet}, LLFlow \cite{llflow}, DCC-Net \cite{dcc-net}, SNR-Net \cite{xu2022snr}, LLFormer \cite{llformer}, PyDiff \cite{zhou2023pyramid}, WCDM \cite{wang2023lldiffusion}, and GSAD \cite{hou2024global}.
\cref{tab:LOL} presents the quantitative results of various LLIE methods, showcasing our method's superior performance across all compared metrics, including PSNR, SSIM, and LPIPS. New SOTA PSNR results of 28.090, 31.250 and 30.166 can be observed on LOLv1, LOLv2-real and LOLv2-synthetic individually. Notably, ReDDiT consistently outperforms other methods, exhibiting substantial improvements in LPIPS scores, indicative of enhanced perceptual quality. Remarkably, ReDDiT achieves SOTA performance on LOLv2-real/LOLv2-synthetic datasets across all distilled models (8, 4, 2 steps). On LOLv1 datasets, our method attains SOTA results with 8 and 4 steps sampling and remains comparable to previous methods with 2 steps.
The visual comparison presented in \cref{fig:visual_compare} further highlights the effectiveness of ReDDiT in mitigating artifacts and enhancing image details. Notably, as depicted in the red box of \cref{fig:visual_compare}, ReDDiT excels in restoring clear structures and intricate details. This underscores our method's ability to leverage generative modeling to capture natural image distributions and retain such characteristics in the distilled models, resulting in superior visual effects.

\noindent \textbf{Results on SID and SDSD.}
 On SID and SDSD, we compare ReDDiT with SID \cite{sid}, DeepUPE \cite{deepupe},   Uformer \cite{wang2022uformer}, RetinexNet \cite{lol1}, KinD \cite{zhang2019kindling},  DRBN \cite{DRBN}, EnlightenGAN \cite{jiang2021enlightengan}, MIRNet \cite{mirnet}, SNR-Net \cite{xu2022snr}, and Retinexformer \cite{cai2023retinexformer}. \cref{tab:SID_SDSD} presents the quantitative results on the SID and SDSD, indicating its capability of ReDDiT to handle more complex low-light conditions effectively.
Specifically, our method establishes new SOTA PSNR values of 25.32 dB/29.95 dB on SID and SDSD dataset. The visual comparison is provided in supplementary materials.

\noindent \textbf{Results on DICM, LIME, MEF, NPE, and VV.} 
We directly apply the model trained on the LOLv2-synthetic dataset to these unpaired real-world datasets. From \cref{tab:unpaired}, it is evident that ReDDiT outperforms all competitors in terms of NIQE scores, demonstrating its robust generalization ability. As shown in \cref{fig:unpair}, our model adeptly adjusts illumination conditions, effectively enhancing visibility in low-light areas while avoiding over-exposure.

\begin{figure*}[t]
\centering
\includegraphics[width=1.0\linewidth]{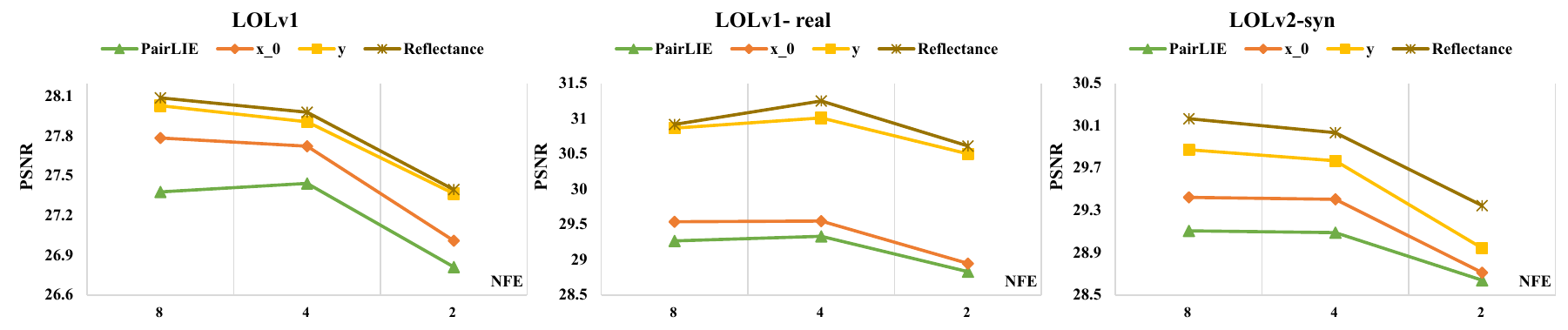} 
\vspace{-0.7cm}
 \caption{An ablation study conducted to evaluate the choice of $\tilde{x}$. The results indicate that reflectance contributes the most to performance during distillation, while low-light images $y$ offer the second-best performance. Both reflectance and low-light images $y$ effectively explore a better residual space, leading to significant improvements in performance compared to the original Gaussian space ($x_0$ as $\tilde{x}$). }
\vspace{-0.5cm}
\label{fig:x_tilde_ablation}
\end{figure*}

\subsection{Comparison with Other Accelerate Methods}
\begin{figure*}[t] 
\centering
\includegraphics[width=1.0\linewidth]{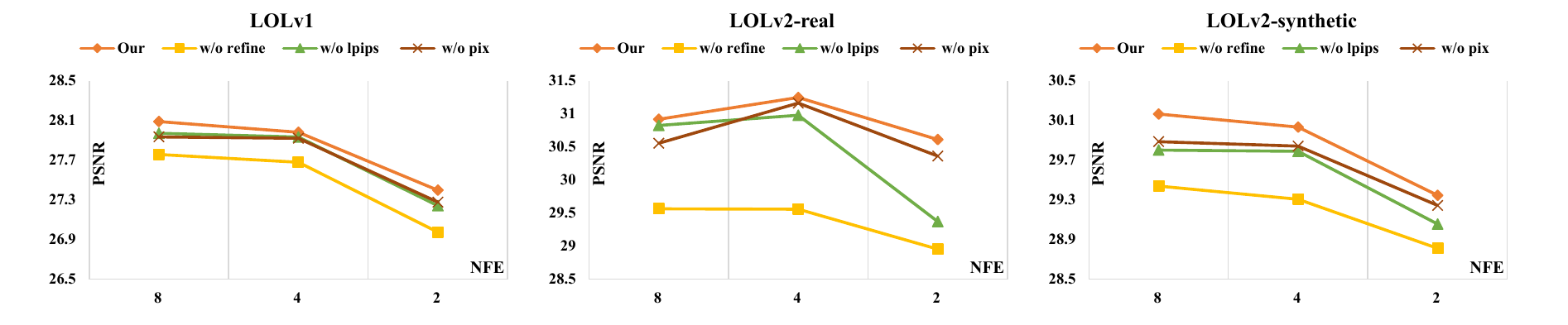} 
\vspace{-0.7cm}
 \caption{The ablation study on the main components of our method. It can be observed that the refinement module significantly contributes to the final performance, with its impact becoming more pronounced as the number of sampling steps decreases.}
\vspace{-0.5cm}
\label{fig:ablation}
\end{figure*}

\begin{figure*}[ht]
\centering
\includegraphics[width=1.0\linewidth]{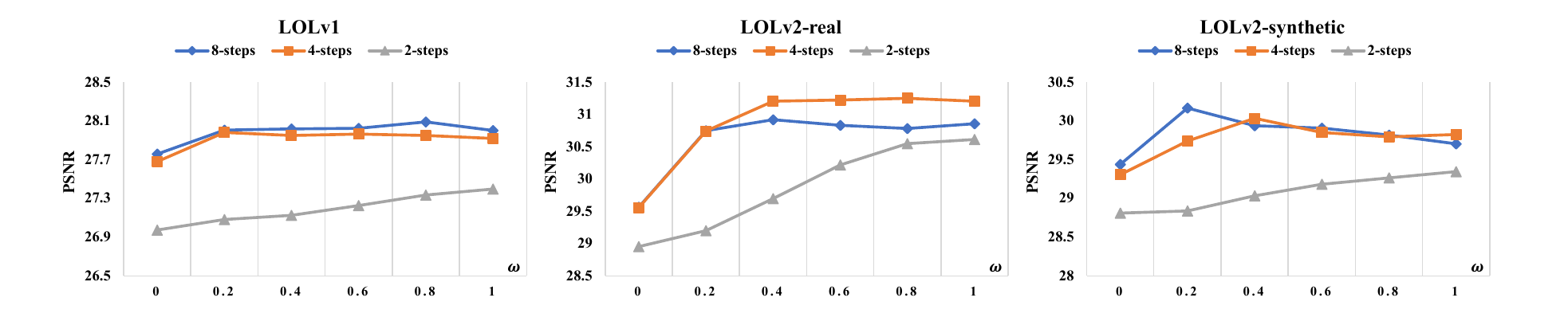} 
\vspace{-0.7cm}
 \caption{The ablation study on the weights of the reflectance-aware trajectory refinement. The horizontal axis represents the strength of the refinement and the vertical axis represents PSNR values.}
\vspace{-0.5cm}
\label{fig:refine_ablation}
\end{figure*}
We conduct a comprehensive evaluation of various acceleration techniques applied to our pre-trained diffusion model. The quantitative results of this comparative analysis are presented in \cref{fig:comparison_with_acc}. Notably, our investigation highlights the critical importance of employing distillation strategies in enhancing model performance.

Among the evaluated techniques, our distillation strategy emerges as the top performer across all datasets and all sampling steps.  
We achieve superior results compared to the teacher model, as demonstrated by the performance improvements over the straightforward application of DDIM. However, it is noteworthy that recent advancements in consistency distillation techniques showcase slight superiority over traditional ODE solver methods, further underscoring the significance of distillation approaches in optimizing model performance.

Most notably, ReDDiT significantly outperforms alternative techniques, demonstrating the substantial benefits of investing in additional training costs to achieve superior results. This emphasizes the pivotal role of distillation strategies in effectively transferring knowledge from a well-prepared teacher model to its student counterpart, thereby elevating the overall performance of ReDDiT. These findings underscore the significance of ReDDiT's superior performance, reinforcing its efficacy and the value of employing sophisticated distillation techniques in advancing the SOTA in low-light image enhancement.

\subsection{Ablation Study}

\textbf{The ablation on refinement module.} The performance degradation without the refinement module is evident in \cref{fig:ablation}, where the absence of this module results in the most significant decline compared to the SOTA performance. Further analysis in \cref{fig:refine_ablation} delves into the impact of the refinement module in more detail. 

The refinement module's influence becomes more pronounced as the number of sampling steps decreases. In the case of 2-step distillation, \cref{fig:refine_ablation} illustrates a consistent performance improvement with increasing strength of the refinements. On LOLv2-real, the 4-step distillation results even surpass the performance of the 8-step diffusion model and achieve new SOTA results with $\omega=0.8$. This trend is attributed to the heightened significance of the refinement module in addressing the increasing fitting error of the teacher model as the number of steps decreases. Therefore, the refinement module becomes crucial for accurately estimating the trajectory under such conditions.

In our previous experiments, we select the refinement strength that yielded the best performance in the previous ablation studies. This decision ensures that the refinement module effectively mitigates the fitting error of the teacher model, thereby enhancing the performance of the student model during distillation.

\noindent \textbf{The ablation on the choice of $\tilde{x}$.} We compare the choice of $\tilde{x}$ with refinement using clean images $x_0$, low-light images $y$, the enhanced result from the learnable Retinex method (\eg PairLIE \cite{pairlie}), and our reflectance component, as shown in \cref{fig:x_tilde_ablation}. Both reflectance and low-light images $y$ effectively explore a better residual space, leading to significant improvements in performance compared to the original Gaussian space ($x_0$ as $\tilde{x}$). This suggests that shifting the residual space is crucial for enhancing the performance of the student model in LLIE tasks.

Furthermore, the PairLIE results, when used for refinement, perform even poorly compared to refinements using clean images $x_0$ and low-light images $y$. Although its prediction is closer to clean images $x_0$, it fails to identify a suitable residual space for the student model to learn from, ultimately failing to address the inference gap.

\noindent \textbf{The ablation on the auxiliary loss.} 
As depicted in \cref{fig:ablation}, conducting ablation studies on $\mathcal{L}_{pix}$ and $\mathcal{L}_{per}$ results in a slight decrease in distillation performance. This observation underscores the effectiveness of directly obtaining supervision signals from ground truth data. Interestingly, $\mathcal{L}_{per}$ exhibits a more pronounced influence on performance compared to $\mathcal{L}_{pix}$, emphasizing the significance of leveraging supervision signals in the feature space.

Furthermore, the degradation in performance resulting from ablation of either loss term is minor compared to the substantial degradation observed when the RATR module is removed. This finding highlights the critical role of the RATR module within our framework. 

\subsection{Efficiency Comparison}
We evaluate the efficiency of our method in terms of inference time, frames per second (FPS), and number of parameters, comparing it with recent diffusion-based LLIE methods. The quantitative results of this comparative analysis are presented in \cref{table:efficiency}. Our approach, particularly in its 2-step variant, demonstrates superior performance regarding inference speed, FPS, and parameter efficiency. It achieves an optimal balance between computational efficiency and model performance, surpassing other methods in both speed and resource utilization. This comparison highlights the effectiveness of our method in delivering high-performance results with minimal computational overhead.

\begin{table}[ht]
\resizebox{0.99\linewidth}{!}{\begin{tabular}{l|c|c|c|c|c|c|c}
\toprule
                    & Teacher-16 & PyDiff & GSAD  & WCDM  & Ours-8 & Ours-4 & Ours-2 \\ \midrule
\textbf{Time (s)} & 0.611      & 0.254  & 0.381 & \underline{0.117} & 0.323  & 0.162  & \bf{0.076}  \\ \midrule
\textbf{FPS}                 & 1.64       & 3.92   & 2.62  & \underline{8.51}  & 3.09   & 6.15   & \bf{13.1}   \\ \midrule
\textbf{Param (M)}      & 17.43      & 97.89  & 17.43 & \underline{22.08} & 17.43  & 17.43  & \bf{17.43} \\ \bottomrule
\end{tabular}}
\vspace{-0.2cm}
\caption{The efficiency comparison with other methods. Our method resulted in a fast and lightweight diffusion-based low-light enhancer with excellent performance.}
\label{table:efficiency}
\vspace{-0.4cm}
\end{table}

\subsection{Limitation and Future work}

While ReDDiT has excelled in restoration within 2 steps, single-step restoration is not yet optimal, presenting a notable limitation. In single-step restoration, inefficient denoising results in low PSNR and artifacts in detail. Failure cases in the single step will be demonstrated in the supplementary materials.
Additionally, the exploration of a lightweight denoising network was not addressed in this work. In our future work, we will continue exploring efficient diffusion-based methods for LLIE. Our focus will be on investigating the potential of a single-step diffusion model and developing a lightweight denoising network. 

\section{Conclusions}
In this paper, we theoretically analyze the two main factors contributing to the performance degradation of the diffusion distillation technique. Building on this, we introduce ReDDiT, a significant advancement in efficient diffusion models for LLIE. Central to ReDDiT is the use of linear exploration in the reflectance-aware residual space, which reduces trajectory fitting errors and the sampling gap. These innovations enable ReDDiT to preserve the intrinsic structural integrity of images while minimizing sampling steps. ReDDiT achieves performance comparable to previous methods with just 2 steps and sets new SOTA results at 4 and 8 steps. Across 10 benchmark datasets, ReDDiT consistently outperforms existing methods. This marks a promising step toward real-time diffusion models for LLIE, and our research in this area will continue.
{   \small
  \bibliographystyle{ieeenat_fullname}
   \bibliography{main}
}


\end{document}